\documentclass[dvipsnames,format=sigconf,authorversion=true]{acmart}

\usepackage{xcolor}
\usepackage{adjustbox}
\usepackage{colortbl}
\usepackage{tabularx}
\usepackage{booktabs} 
\usepackage{graphicx}
\usepackage{subfig}
\usepackage[linesnumbered, vlined, ruled]{algorithm2e}
\SetArgSty{textup}
\usepackage{enumerate}
\usepackage{prettyref}
\usepackage{color, colortbl}
\usepackage{braket}
\definecolor{MyHiLiRow}{gray}{0.9}
\usepackage{tcolorbox}
\usepackage{caption}
\usepackage{seqsplit}
\usepackage{multirow}
\usepackage{tcolorbox}
\usepackage{adjustbox}

\def\vec#1{\mathbf{#1}}
\def\set#1{\mathcal{#1}}

\newcommand{\pref}{\prettyref}

\newrefformat{fig}{Figure~\ref{#1}}
\newrefformat{supfig}{Figure~\ref{#1}}
\newrefformat{tab}{Table~\ref{#1}}
\newrefformat{suptab}{Figure~\ref{#1}}
\newrefformat{sec}{Section~\ref{#1}}
\newrefformat{subsection}{Section~\ref{#1}}
\newrefformat{subsec}{Section~\ref{#1}}
\newrefformat{app}{Appendix~\ref{#1}}
\newrefformat{alg}{Algorithm~\ref{#1}}
\newrefformat{property}{Property~\ref{#1}}
\newrefformat{theorem}{Theorem~\ref{#1}}
\newrefformat{corollary}{Corollary~\ref{#1}}
\newrefformat{proposition}{Proposition~\ref{#1}}
\newrefformat{def}{Definition~\ref{#1}}
\newrefformat{eq}{equation~(\ref{#1})}

\AtBeginDocument{%
  \providecommand\BibTeX{{%
    \normalfont B\kern-0.5em{\scshape i\kern-0.25em b}\kern-0.8em\TeX}}}

\copyrightyear{2024}
\acmYear{2024}
\setcopyright{acmlicensed}\acmConference[GECCO '24 Companion]{Genetic and Evolutionary Computation Conference}{July 14--18, 2024}{Melbourne, VIC, Australia}
\acmBooktitle{Genetic and Evolutionary Computation Conference (GECCO '24 Companion), July 14--18, 2024, Melbourne, VIC, Australia}
\acmDOI{10.1145/3638530.3664127}
\acmISBN{979-8-4007-0495-6/24/07}

\begin{document}

\title[On Constructing Algorithm Portfolios in Algorithm Selection for Computationally Expensive BBO]{On Constructing Algorithm Portfolios in Algorithm Selection for Computationally Expensive Black-box Optimization in the Fixed-budget Setting}





\author{Takushi Yoshikawa}
\affiliation{%
  \institution{Yokohama National University}
  \city{Yokohama}
  \state{Kanagawa}
  \country{Japan}}
  \email{yoshikawa-takushi-nj@ynu.jp}
  
\author{Ryoji Tanabe}
\affiliation{%
  \institution{Yokohama National University}
  \city{Yokohama}
  \state{Kanagawa}
  \country{Japan}}
  \email{rt.ryoji.tanabe@gmail.com}







\begin{abstract}

Feature-based offline algorithm selection has shown its effectiveness in a wide range of optimization problems, including the black-box optimization problem.
An algorithm selection system selects the most promising optimizer from an algorithm portfolio, which is a set of pre-defined optimizers. 
Thus, algorithm selection requires a well-constructed algorithm portfolio consisting of efficient optimizers complementary to each other.
Although construction methods for the fixed-target setting have been well studied, those for the fixed-budget setting have received less attention.
Here, the fixed-budget setting is generally used for computationally expensive optimization, where a budget of function evaluations is small.
In this context, first, this paper points out some undesirable properties of experimental setups in previous studies.
Then, this paper argues the importance of considering the number of function evaluations used in the sampling phase when constructing algorithm portfolios, whereas the previous studies ignored that.
The results show that algorithm portfolios constructed by our approach perform significantly better than those by the previous approach.


\end{abstract}





\maketitle

\section{Introduction}
\label{sec:introduction}


This paper considers a single-objective noiseless black-box optimization of an objective function $f: \Omega \rightarrow \mathbb{R}$, where $\Omega \subseteq \mathbb{R}^n$ is the $n$-dimensional solution space.
This problem aims to find a solution $\vec{x} \in \Omega$ with an objective value $f(\vec{x})$ as small as possible without any explicit knowledge of $f$.
%
A number of black-box derivative-free optimizers have been proposed, including evolutionary algorithms.

In general, the best optimizer significantly depends on the property of a given function~\cite{HansenARFP10}.
Representative properties include multimodality, global multimodality, variable separability, and ill-condition.
Therefore, in practical applications, the user needs to select the most promising one from multiple candidate optimizers for the target function.
However, this hand-selecting requires knowledge of optimizers and tedious trial-and-error.

Automated algorithm selection~\cite{Rice76,KerschkeHNT19} can potentially address this issue.
Automated algorithm selection automatically selects the most promising one from multiple candidate optimizers in a human-out-of-the-loop manner.
Given an algorithm portfolio $\set{A}=\{a_1, \ldots, a_k\}$ of size $k$, a set of target function instances $\set{I}$, and a performance measure $m:\set{A} \times \set{I} \rightarrow \mathbb{R}$, the algorithm selection problem involves selecting the best optimizer $a_{\mathrm{best}}$ in terms of $m$~\cite{Rice76,LindauerRK19}.
The algorithm selection problem frequently appears in a wide range of real-world applications, including the aforementioned black-box optimization problem.

Feature-based off-line algorithm selection is a useful approach to the algorithm selection problem~\cite{KerschkeHNT19}.
In the feature-based off-line automatic algorithm selection, features are first computed on the target function based on a small-sized solution set $\set{X}$.
Each feature should represent one or more properties of the target function.
Then, an algorithm selection system predicts the best optimizer $a_{\mathrm{best}}$ from the algorithm portfolio $\set{A}$ based on the feature set.
A machine learning model is generally used for this prediction.
Finally, the selected optimizer is presented to the user.
Feature-based off-line algorithm selection has demonstrated its effectiveness on a wide range of search and optimization problems, including SAT~\cite{XuHHL08} and TSP~\cite{KerschkeKBHT18}.

Recently, algorithm selection has received much attention in the field of black box optimization ~\cite{KerschkeT19,DerbelLVAT19,BischlMTP12,JankovicPED21,tanabe2022benchmarking}.
In black-box optimization, the features are generally computed using the exploratory landscape analysis (ELA)~\cite{MersmannBTPWR11}.
First, a solution set $\set{X}$ of size $s$ is randomly sampled, and their corresponding objective function values $f(\set{X})$ are computed.
ELA features are computed based on the pairs of $\set{X}$ and $f(\set{X})$.

The choice of optimizers in an algorithm portfolio $\set{A}$ significantly influences the performance of algorithm selection systems \cite{tanabe2022benchmarking}.
It is desirable that an algorithm portfolio is as small as possible and consists of efficient optimizers complementary to each other~\cite{KerschkeT19}. 
As reviewed in \cite{tanabe2022benchmarking}, some methods for automatically constructing algorithm portfolios \cite{KerschkeT19,BischlMTP12,MunozK16} have been proposed for black-box optimization.

Fixed-target and fixed-budget settings are available for benchmarking black-box optimizers~\cite{HansenABTT16,HansenABT22}.
Let $f^{\mathrm{target}}$ be the pre-defined objective value of a solution to be reached.
In the fixed-target setting, the performance of optimizers is evaluated based on the number of function evaluations used to reach $f^{\mathrm{target}}$.
In contrast, in the fixed-budget setting, the performance of optimizers is evaluated based on how good the quality of the best-so-far solution is obtained within a pre-defined budget of function evaluations.
As discussed in \cite{HansenABTT16,HansenABT22}, it is preferable to use the fixed-target setting for benchmarking black-box optimizers.


The computational cost of the objective function is high in some real-world applications of black-box optimization.
For example, the computation time to evaluate a single solution by the objective function is about one minute for the robot controller design problem~\cite{rebolledo2020parallelized} and about one hour for the car body design problem~\cite{le2013evolution}.
In such a case, the maximum number of function evaluations available to the optimizers must be strictly limited (e.g., $100 \times n$) to reduce the computational time for optimization.
A number of efficient approaches have been proposed for computationally expensive black-box optimization problems, including Bayesian optimization and surrogate-assisted evolutionary algorithms~\cite{WangJSO23}.

However, most previous studies on feature-based off-line algorithm selection for black-box optimization (e.g., \cite{KerschkeT19,DerbelLVAT19,BischlMTP12,JankovicPED21,tanabe2022benchmarking}) did not address computationally expensive optimization.
Most previous studies also considered the fixed-target setting.
However, it is difficult to set the target objective value $f^{\mathrm{target}}$ to a suitable value on multiple test functions for computationally expensive optimization.
This is because most optimizers cannot find good solutions when the maximum number of function evaluations is strictly limited.
Therefore, the fixed-budget setting can only be used for computationally expensive optimization. 
Some previous studies (e.g., \cite{JankovicD20,JankovicPED21}) used the fixed-budget setting and set the maximum number of function evaluations to a small number.
However, as described in \pref{sec:proposed_method}, some experimental settings in \cite{JankovicD20,JankovicPED21} are unrealistic.

Most previous studies (e.g., \cite{KerschkeT19,DerbelLVAT19,tanabe2022benchmarking}) set the size $s$ of the solution set $\set{X}$ used in the ELA feature computation to about $50 \times n$.
However, $s$ must be set to a smaller value when the maximum number of evaluations is strictly limited.
Otherwise, the budget of function evaluations available to an optimizer selected by algorithm selection can be exhausted.
However, the setting of $s$ to a too small value can degrade the effectiveness of ELA features~\cite{KerschkePWT15}.

How to construct algorithm portfolios for the fixed-budget setting has not been discussed in the literature.
Here, all the four existing construction methods \cite{KerschkeT19,BischlMTP12,MunozK16,tanabe2022benchmarking} were designed for the fixed-target setting.
Let \texttt{MaxFE} be the maximum number of function evaluations for \emph{algorithm selection systems (not optimizers)}.
Generally, algorithm portfolios are constructed based on the performance of optimizers until \texttt{MaxFE}. 
However, the actual maximum number of function evaluations for \emph{optimizers (not algorithm selection systems)} is $\texttt{MaxFE} - s$.
On the one hand, the difference between \texttt{MaxFE} and $\texttt{MaxFE} - s$ may not be problematic for the fixed-target scenario and the fixed-budget scenario using a large \texttt{MaxFE}.
This is because the influence of $s$ is negligible.

On the other hand, the influence of $s$ is not negligible when \texttt{MaxFE} is limited to a small number, i.e., computationally expensive optimization.
For example, let us consider that $\texttt{MaxFE} = 100n$ and $s = 50n$.
In this case, the generation of the solution set $\set{X}$ for the ELA feature computation requires $50n$ function evaluations, and an optimizer selected by algorithm selection can use only the remaining $50n$ function evaluations.
If an algorithm portfolio is constructed based on the performance of optimizers until $100n$ function evaluations, the resulting algorithm portfolio is likely to include a poorly performing optimizer at $50n$ function evaluations.



This paper investigates how to construct algorithm portfolios in feature-based automatic algorithm selection for computationally expensive black-box optimization.
Throughout this paper, we address the fixed-budget setting and computationally expensive black-box optimization.
\pref{sec:preliminary} gives some preliminaries.
In \pref{sec:proposed_method}, we point out some issues of the experimental settings in the previous studies \cite{JankovicD20,JankovicPED21}.
Then, we describe our approach.
\pref{sec:setup} describes the experimental setup.
In \pref{sec:results}, we address the following two research questions (RQ) through experimental analysis:

\begin{enumerate}[RQ1:]
\item How does the difference between $\texttt{MaxFE}$ and $\texttt{MaxFE} - s$ influence the effectiveness of resulting algorithm portfolios?
\item Can algorithm selection systems outperform the single-best solver (SBS)?
\end{enumerate}

\noindent Finally, \pref{sec:conclusion} concludes this paper.
\section{Preliminaries}
\label{sec:preliminary}

This paper uses the terminologies defined in \cite{HansenARMTB21}.
An objective \underline{function} performs a parameterized mapping $\mathbb{R}^n \rightarrow \mathbb{R}$.
A function \underline{instance} is an instantiated version of the function by giving parameters.
For example, $f_1$ (the Sphere function) in the noiseless BBOB function set ~\cite{HansenFRA12} is defined as follows: $f(\vec{x}) = \|\vec{x} - \vec{x}_{\mathrm{opt}}\| + f_{\mathrm{opt}}$.
Here, $f_1$ is said to be a \emph{function}.
In contrast, $f_1$ with $n=2$, $\vec{x}_{\mathrm{opt}}=(1, 2)^{\top}$, and $f_{\mathrm{opt}}=-100$ is said to be a function \emph{instance}.
A \underline{problem} is a function instance to which an optimizer is applied.
However, a target objective value $f^{\mathrm{target}}$ can be considered to define the problem.

\subsection{Automatic algorithm selection}
\label{subsec:as}

Feature-based automatic algorithm selection consists of a training phase and a testing phase.
Given a portfolio $\set{A}=\{a_j\}^{k}_{j=1}$ of size $k$ and a training instance set $\set{I}^{\mathrm{train}}$, we assume that the performance of each optimizer $a \in \set{A}$ on each instance $i \in \set{I}^{\mathrm{train}}$ is known.

In the training phase, for each instance $i \in \set{I}^{\mathrm{train}}$, a solution set $\set{X}=\{\vec{x}_j\}^{s}_{j=1}$ of size $s$ is generated by a sampling method, e.g., Latin hypercube sampling.
Then, the quality of $\set{X}$ is evaluated by the objective function $f$ to obtain the corresponding objective function value set $f(\set{X})$.
This pair of $\set{X}$ and $f(\set{X})$ is used to compute the feature set $\set{F}$ (see \pref{subsec:features}), which is used to train a machine learning model.
Here, the trained machine learning model is used for algorithm selection.

In the testing phase, a test instance set $\set{I}^{\mathrm{test}}$ is given, where the performance of each optimizer $a \in \set{A}$ on $i \in \set{I}^{\mathrm{test}}$ is unknown.
Here, $|\set{I}^{\mathrm{test}}|=1$ in most real-world applications.
For each test instance $i \in \set{I}^{\mathrm{test}}$, the feature set $\set{F}$ is first calculated in the same way as in the training phase.
Then, the performance of each optimizer on $i \in \set{I}^{\mathrm{test}}$ is predicted by the trained machine learning model.
Finally, the optimizer with the best prediction performance is actually applied to $i \in \set{I}^{\mathrm{test}}$.


Although some selection methods have been proposed in the literature, the previous study~\cite{tanabe2022benchmarking} showed that the regression-based selection method \cite{XuHHL08} performs well.
In the training phase, the regression-based selection method constructs $k$ regression models based on $\set{F}$ for the $k$ optimizers $a_1, \ldots, a_k$.
Here, for $j \in \{1, \ldots, k\}$, the $j$-th regression model predicts the performance of the $j$-th optimizer on $\set{I}^{\mathrm{train}}$.
In the testing phase, for each $i \in \set{I}^{\mathrm{test}}$, the optimizer with the best prediction performance is selected from the $k$ optimizers $a_1, \ldots, a_k$.

The virtual best solver (VBS) and the single best solver (SBS) play an important role in benchmarking algorithm selection systems.
The VBS is the ideal algorithm selection system that always selects the best optimizer from the algorithm portfolio $\set{A}$.
The SBS is the best single optimizer in $\set{A}$ on all functions considered in terms of a performance measure $m$.
The VBS represents the upper bound of the performance of algorithm selection systems.
In contrast, the SBS represents the performance of algorithm selection systems that should be shown even in the worst case.
If an algorithm selection system using $\set{A}$ is outperformed by the SBS in $\set{A}$, it means that the algorithm selection system does not work well.

\subsection{ELA features}
\label{subsec:features}

Although some feature computation approaches for black-box optimization have been proposed in previous studies, exploratory landscape analysis (ELA)~\cite{MersmannBTPWR11} is the most representative one.
ELA computes the feature set $\set{F}$ based on the solution set $\set{X}$ of size $s$ and the corresponding objective function value set $f(\set{X})$ described in \pref{subsec:as}.
If $s$ is set to a large number, the actual maximum number of function evaluations available for the selected optimizer becomes small.
Thus, $s$ should be set as small as possible.
However, the usefulness of the ELA features decreases when setting $s$ to too small.


Most previous studies computed the ELA features by using the software \texttt{flacco}~\cite{KerschkeT2019flacco} implemented in \textsf{R}.
\pref{tab:flacco_features} shows the nine ELA feature classes provided by \texttt{flacco}.
The $\texttt{ela\_conv}$, $\texttt{ela\_curv}$, and $\texttt{ela\_local}$ feature classes require additional function evaluations.
In addition, feature classes such as $\mathtt{cm\_angle}$ can be computed only when $n \leq 5$.
For this reason, these feature classes have not been used in many previous studies.
Therefore, we do not consider these features in this work as well.

Each feature class aims to quantify one or more function properties.
Each feature class also consists of one or more features.
For example, $\texttt{ela\_level}$ consists of 20 features based on the distribution of the objective function value set $f(\set{X})$. 
In $\texttt{ela\_level}$, pairs of $\set{X}$ and $f(\set{X})$ are classified into two classes based on predefined threshold values.
Then, the three classification models are applied to the classification problem for each objective function value of $f(\set{X})$.
The result can be each feature in $\texttt{ela\_level}$.
For example, the $\texttt{ela\_level.mmce\_lda\_10}$ feature in the $\texttt{ela\_level}$ class is the average error value obtained by applying LDA with the threshold as the upper $10\%$ of $f(\set{X})$.

\begin{table}[t]
\centering
\caption{Nine feature classes provided by \texttt{flacco}~\cite{KerschkeT2019flacco}.}
\label{tab:flacco_features}
\renewcommand{\arraystretch}{0.85}  
{\small
\scalebox{1}[1]{ 
\begin{tabular}{llc}
  \toprule
Feature class & Name & Num. features\\  
\midrule
\texttt{ela\_distr}~\cite{MersmannBTPWR11} & $y$-distribution & 5\\
\texttt{ela\_level}~\cite{MersmannBTPWR11} & levelset & 20\\
\texttt{ela\_meta}~\cite{MersmannBTPWR11} & meta-model & 11\\
\texttt{nbc}~\cite{KerschkePWT15} & nearest better clustering & 7\\
\texttt{disp}~\cite{KerschkePWT15} & dispersion & 18\\
\texttt{ic}~\cite{MunozKH15} & information content & 7\\
\texttt{basic}~\cite{KerschkeT2019flacco} & basic & 15\\
\texttt{limo}~\cite{KerschkeT2019flacco} & linear model & 14\\
\texttt{pca}~\cite{KerschkeT2019flacco} & principal component analysis & 10\\
\bottomrule
\end{tabular}
}
}
\end{table}

\subsection{Cross-validation}
\label{subsec:cross_validation}

Most previous studies evaluated the performance of algorithm selection systems using the noiseless BBOB function set ~\cite{HansenFRA12}.
Each BBOB function represents at least one difficulty found in real-world problems.
Here, the BBOB function set consists of 24 various test functions $f_1, \ldots, f_{24}$.
Moreover, each test function consists of countless instances, which differ in the location of the optimal solution, the elements of the rotation matrix used for the axis transformation, and so on.

As described in \pref{subsec:as}, automatic algorithm selection requires a training instance set $\set{I}^{\mathrm{train}}$.
However, $\set{I}^{\mathrm{train}}$ should be different from the test instance set $\set{I}^{\mathrm{test}}$.
In the following, we introduce two representative cross-validation methods: leave-one-instance-out cross-validation (LOIO-CV)~\cite{BischlMTP12,JankovicPED21} and leave-one-function-out cross-validation (LOFO-CV)~\cite{BischlMTP12,DerbelLVAT19}.
Here, the number of instances of the $j$-th BBOB function $f_j$ is 5 for each $j \in \{1, \dots, 24\}$.
We also denote the set of five instances of $f_j$ by $\set{I}_j$.

In the LOIO-CV, 5-fold cross-validation is performed for each dimension.
For each $j \in \{1, \ldots, 5\}$, $\set{I}^{\mathrm{test}}$ in the $j$-th fold is the set of $j$-th instances of the 24 functions. 
That is, $|\set{I}^{\mathrm{test}}|=24 \times 1 = 24$.
Therefore, $\set{I}^{\mathrm{train}}$ is the set of the remaining $24 \times 4 = 96$ instances.
Since $\set{I}^{\mathrm{test}}$ and $\set{I}^{\mathrm{train}}$ always contain instances of the same function, $\set{I}^{\mathrm{test}}$ and $\set{I}^{\mathrm{train}}$ are very similar.
As demonstrated in \cite{tanabe2022benchmarking}, algorithm selection systems can outperform the SBS in most cases when using the LOIO-CV.
In other words, the LOIO-CV is easy for algorithm selection systems.

In the LOFO-CV, 24-fold cross-validation is performed for each dimension.
For each $j \in \{1, \ldots, 24\}$, $\set{I}^{\mathrm{test}}$ in the $j$-th fold is a set of 5 instances of the $j$-th function, i.e., $\set{I}^{\mathrm{test}}=\set{I}_j$.
That is, $|\set{I}^{\mathrm{test}}|=1 \times 5 = 5$.
$\set{I}^{\mathrm{train}}$ is the set of the remaining $23 \times 5 = 115$ instances.
Unlike the LOIO-CV, the test and training instance sets are very different in the LOFO-CV.
Note that the 24 BBOB functions have different properties from each other.
Therefore, the LOFO-CV is more challenging than the LOIO-CV~\cite{tanabe2022benchmarking}.
Problem instances in the testing and training phases should be different when evaluating the generalization performance of algorithm selection systems.
In fact, the properties of a real-world problem can be dissimilar to those of the 24 BBOB functions~\cite{SteinWB20}. 
For this reason, the previous study~\cite{tanabe2022benchmarking} recommended using the LOFO-CV.


\begin{table*}[t]  
\centering
\caption{Experimental settings in the previous study~\cite{JankovicD20} and this paper. ``Consideration of $\set{X}$" indicates whether the number of $s$ function evaluations spent on evaluating the solution set $\set{X}$ of size $s$ is considered or not. The $\texttt{MaxFE}$ denotes the maximum number of function evaluations described in each paper.}
{\small
  \label{tab:setup}
\begin{tabular}{lcccccccc}
\toprule 
& Cross-validation & $n$ & Consideration of $\set{X}$ & $s$ & \texttt{MaxFE} & Actual \texttt{MaxFE} & \texttt{MaxFE} for an optimizer\\
\midrule
\multirow{2}{*}{The previous study~\cite{JankovicD20}} & \multirow{2}{*}{LOIO-CV} & \multirow{2}{*}{$5$} & & $400n$ & $50n$ &  $\texttt{MaxFE} + s$ & $\texttt{MaxFE}$\\
 & & & & $50n$ & $50n$ &  $\texttt{MaxFE} + s$ &  $\texttt{MaxFE}$\\
 \midrule
\raisebox{0.5em}{This paper} & \raisebox{0.5em}{LOFO-CV} & \shortstack{$2, 3,$\\$ 5, 10$} & \raisebox{0.5em}{Yes} & \shortstack{$10n, 15n, 20n,$\\$25, 50n$} & \raisebox{0.5em}{$100n$} & \raisebox{0.5em}{\texttt{MaxFE}} & \raisebox{0.5em}{$\texttt{MaxFE} - s$}\\
\bottomrule 
\end{tabular}
}
\end{table*}

\section{Some issues in previous studies and our approach}
\label{sec:proposed_method}

As described in \pref{sec:introduction}, we consider the fixed-budget setting.
We notice the disadvantage of the fixed-budget setting, i.e., it is difficult to quantitatively discuss the performance of optimizers.
However, it is difficult to determine the target objective function value $f_{\mathrm{target}}$ for the fixed-target setting when the maximum number of function evaluations is strictly limited.
The two previous studies~\cite{JankovicD20,JankovicPED21} addressed the fixed-budget setting.
However, the experimental settings in \cite{JankovicD20,JankovicPED21} have some issues.

\pref{tab:setup} shows the experimental settings in  the previous studies~\cite{JankovicD20} and this work.
First, \pref{subsection:prev_studies} points out the issues in the previous study~\cite{JankovicD20}.
Since the experimental settings in \cite{JankovicD20,JankovicPED21} are almost the same, we describe only those in \cite{JankovicD20}.
Then, \pref{subsection:our_study} describes our approach based on the discussion in \pref{subsection:prev_studies}.

\subsection{Issues in previous studies}
\label{subsection:prev_studies}

As shown in \pref{tab:setup}, the previous study ~\cite{JankovicD20} used the LOIO-CV for cross-validation.
However, as mentioned in \pref{subsec:cross_validation}, the LOIO-CV is not recommended for benchmarking algorithm selection systems.
The previous study ~\cite{JankovicD20} did not consider the scalability of the algorithm selection system with respect to $n$.
As discussed in \cite{tanabe2022benchmarking}, the number of function evaluations $s$ spent in evaluating the solution set $\set{X}$ should be considered as the number of evaluations spent in the whole algorithm selection system.
Here, $\set{X}$ is used for the feature computation.
Since the previous study ~\cite{JankovicD20} does not take this into account, the comparison is not realistic.
In the main experiments in \cite{JankovicD20}, the size $s$ of the solution set $\set{X}$ was set to $400n$.
Here, the maximum number of function evaluations (\texttt{MaxFE}) was set to $50n$.
Thus, $s$ is larger than \texttt{MaxFE}.
We point out that the actual \texttt{MaxFE} used in \cite{JankovicD20} should be $\texttt{MaxFE} + s = 50n + 400n = 450n$.
In \cite{JankovicD20}, the maximum number of function evaluations for the selected optimizer is set to \texttt{MaxFE}.
However, as discussed in \pref{sec:introduction}, this setting causes a contradiction between the predicted and actual performance of optimizers.



\subsection{Our approach}
\label{subsection:our_study}

Based on the discussion in \pref{subsection:prev_studies}, we use the more challenging and practical LOFO-CV for cross-validation.
We also investigate the scalability of algorithm selection systems with respect to $n$.
We consider the number of $s$ function evaluations for generating $\set{X}$.
We investigate the influence of $s$ on the effectiveness of algorithm selection systems.
Thus, we set $s$ to $10n, 15n, 20n, 25n, $ and $50n$.
Here, as far as we know, no previous study set $s$ to $10n$.
We set \texttt{MaxFE} to $100n$, which has been often used in the computationally expensive optimization setting in the BBOB function suite.

While the actual \texttt{MaxFE} available for algorithm selection systems in the previous study~\cite{JankovicD20} was $\texttt{MaxFE} + s$ ($=50n + 400n = 450n$), that in our study is exactly \texttt{MaxFE} ($100n$).
Thus, the maximum number of function evaluations available for a selected optimizer is $\texttt{MaxFE} - s$.
Taking into account this fact, we investigate the effectiveness of algorithm portfolios constructed by referring to the performance of optimizers until $\texttt{MaxFE} - s$, where the previous study~\cite{JankovicD20} referred to the performance of optimizers until $\texttt{MaxFE}$.

\definecolor{c1}{RGB}{150,150,150}
\definecolor{c2}{RGB}{220,220,220}

\section{Experimental setup}
\label{sec:setup}

In this study, we used a workstation with a 40-core Intel(R) Xeon Gold 6230 (20 cores $\times$ 2CPU) 2.7GHz with 384GB RAM using Ubuntu 22.04.
Based on the suggestion by \cite{tanabe2022benchmarking}, we performed 31 independent trials of algorithm selection systems.
We used the noiseless BBOB function set ~\cite{HansenFRA12}.
We used benchmarking data of 244 optimizers provided in the COCO archive (\url{https://numbbo.github.io/data-archive/bbob/}).
We set $n$ to $2, 3, 5, $ and $10$.
As in the previous study ~\cite{KerschkeT19,tanabe2022benchmarking}, we set the number of instances of each BBOB function to 5.

We consider the fixed-budget setting.
We evaluated the performance of each optimizer based on the objective value $f(\vec{x}^{\mathrm{bsf}})$ of the best-so-far solution $\vec{x}^{\mathrm{bsf}}$ found until the pre-defined budget of function evaluations.
We used the error value $|f(\vec{x}^{\mathrm{bsf}}) - f(\vec{x}^*)|$ between  $f(\vec{x}^{\mathrm{bsf}})$ and the optimal value $f(\vec{x}^*)$ as the performance measure for each optimizer. 
As mentioned in \pref{subsection:our_study}, we set the maximum number of evaluations to $100n$.

We set the size $s$ of the solution set $\set{X}$ used for the ELA feature computation to $10n, 15n, 20n, 25n, $ and $50n$.
As mentioned in the \pref{subsection:our_study}, the maximum number of evaluations available for an optimizer selected by an algorithm selection system is $100n-s$, i.e., $90n$ for $s=10n$, $85n$ for $s=15n$, $80n$ for $s=20n$, $75n$ for $s=25n$, and $50n$ for $s=50n$.
In this work, we used the ELA features shown in \pref{tab:flacco_features}.
We used the \textsf{R} software \texttt{flacco}~\cite{KerschkeT2019flacco} for the feature computation. 
Technically, we used an early version of \texttt{pflacco} (version $0.4$)~\cite{PragerT24}, which provides the Python interface of \texttt{flacco}.
However, in our preliminary experiment, the \texttt{nbc} feature class could not be computed for $s=10n$.
Therefore, we did not use the $\mathtt{nbc}$ feature class only for $s=10n$.
We imputed missing features using the average value of the same features for training. 

We used the LOFO-CV for cross-validation.
This work used the regression-based selection method described in \pref{subsec:as} for algorithm selection.
The previous study~\cite{tanabe2022benchmarking} reported that the regression-based selection method performs better for the LOFO-CV than other selection methods. 
We used the off-the-shelf random forest with the default parameters implemented in scikit-learn.

\subsection{Algorithm portfolios}
\label{subsec:portfolio}

As in \cite{tanabe2022benchmarking}, we used the local search method for the general subset selection problem~\cite{BasseurDGL16} to construct algorithm portfolios.
One advantage of this approach is that it can determine the size $k$ of an algorithm portfolio $\set{A}$.

In this local search approach, first, the 244 optimizers are ranked based on the performance measure, where we used the error value $|f(\vec{x}^{\mathrm{bsf}}) - f(\vec{x}^*)|$ as mentioned above.
Then, local search selects $k$ optimizers from the 244 optimizers so that the sum of the rankings of the $k$ optimizers is minimized as possible.
As in \cite{BischlMTP12}, we set $k$ to 4 in this study.

We ranked the 244 optimizers for each BBOB function and each $n$.
The optimizers are sorted in ascending order based on the average of the error values on the five function instances.
When multiple optimizers achieved the same average error value, we assigned the same rank to them, i.e., we did not use any tie-breaker.
Here, multiple optimizers in each algorithm portfolio found the optimal solution on some easy-to-solve functions (i.e., the Sphere function $f_1$ and the linear slop function $f_5$).

\pref{tab:alg_portfolio} shows six algorithm portfolios $\mathcal{A}_0, \ldots, \mathcal{A}_{50}$ constructed in this work.
In \pref{tab:alg_portfolio}, $\set{A}_0$ is an algorithm portfolio constructed without considering the budget of $s$ function evaluations for generating the solution set $\set{X}$ of size $s$.
Thus, $\set{A}_0$ is an algorithm portfolio constructed based on the performance of the 244 optimizers until the maximum number of evaluations $100n$.
This approach is the same as in the previous study ~\cite{JankovicD20} described in \pref{sec:proposed_method}.
Except for $\set{A}_0$, we denote an algorithm portfolio for $s$ as $\set{A}_s$.
Thus, for $s \in \{10n, 15n, 20n, 25n, 50n\}$, $\set{A}_{10}$, $\set{A}_{15}$, $\set{A}_{20}$, $\set{A}_{25}$, and $\set{A}_{50}$ are constructed based on the performance of optimizers until $90n$, $85n$, $80n$, $75n$, and $50n$ function evaluations.
In each portfolio, the SBS is highlighted with {\adjustbox{margin=0.1em, bgcolor=c1}{dark gray}.
The optimizers selected in \pref{tab:alg_portfolio} can be roughly classified into the following five groups: (a) conventional CMA-ES, (b) surrogate model-based CMA-ES, (c) mathematical derivative-free optimizers, and (d) DIRECT-based optimizers, and (e) STEP-based optimizers that were designed for separable functions~\cite{LangermanSB94}.
In any algorithm portfolio in \pref{tab:alg_portfolio}, the surrogate model-based CMA-ES (lq-CMA-ES~\cite{Hansen19} and DTS-CMA-ES~\cite{BajerPRH19}) is the SBS.




\begin{table}[t]
\centering
\caption{Six algorithm portfolios of size $k=4$ constructed in this work. Symbols (a)--(e) denote the classification of each optimizer. For details, see the paper.
}
\label{tab:alg_portfolio}
{\normalsize
\begin{tabular}{ll}
\toprule
$\mathcal{A}$ & Four optimizers in $\mathcal{A}$\\
\midrule
$\mathcal{A}_0$ & oMads-2N (d), MLSL (c), \\ & \adjustbox{margin=0.1em, bgcolor=c1}{lq-CMA-ES} (b), BIPOP-aCMA-STEP (e)\\
\midrule
$\mathcal{A}_{10}$ & BrentSTEPif (e), CMA-ES-2019 (a), \\ & DIRECT-REV  (d), \\ & \adjustbox{margin=0.1em, bgcolor=c1}{DTS-CMA-ES$\_$005-2pop$\_$v26$\_$1model} (b)\\
\midrule
$\mathcal{A}_{15}$ & lmm-CMA-ES (b), \adjustbox{margin=0.1em, bgcolor=c1}{lq-CMA-ES} (b), \\ & STEPifeg (e), fmincon (c)\\
\midrule
$\mathcal{A}_{20}$ & lmm-CMA-ES (b), \adjustbox{margin=0.1em, bgcolor=c1}{lq-CMA-ES} (b), \\ & STEPifeg (e), fmincon (c)\\
\midrule
$\mathcal{A}_{25}$ & lmm-CMA-ES (b), \adjustbox{margin=0.1em, bgcolor=c1}{lq-CMA-ES} (b), \\ & DIRECT-REV (d), BrentSTEPif (e)\\
\midrule
$\mathcal{A}_{50}$ & lmm-CMA-ES (b), \adjustbox{margin=0.1em, bgcolor=c1}{lq-CMA-ES} (b), \\ & BrentSTEPrr (e), oMads-2N (d)\\
\bottomrule
\end{tabular}}\\
\end{table}

\definecolor{c1}{RGB}{150,150,150}
\definecolor{c2}{RGB}{220,220,220}

\section{Results}
\label{sec:results}

Through experimental analysis, \pref{subsec:rq1} and \pref{subsec:rq2} address RQ1 and RQ2 described in \pref{sec:introduction}, respectively.

\subsection{The influence of algorithm portfolios on the performance of algorithm selection systems}
\label{subsec:rq1}

\subsubsection{Comparisons of algorithm selection systems}

\pref{tab:friedman_ranking} shows the comparison of 10 algorithm selection systems using the six algorithm portfolios in \pref{tab:alg_portfolio} on the noiseless BBOB function suite for each $n$.
To investigate the influence of $s$, we evaluated the performance of the algorithm selection system with $\set{A}_{0}$ using $s = 10n, 15n, 20n, 25n,$ and $50n$.
Therefore, we consider the 10 algorithm selection systems in total.
For $\set{A}_{10}, \ldots, \set{A}_{50}$, we used the corresponding $s$.
\pref{tab:friedman_ranking} shows the Friedman test-based average rankings of the 10 algorithm selection systems.
We used the CONTROLTEST software~\cite{GarciaFLH10} (\url{https://sci2s.ugr.es/sicidm}) to obtain the rankings.
In \pref{tab:friedman_ranking}, 
The best and second-best data are highlighted in {\adjustbox{margin=0.1em, bgcolor=c1}{dark gray} and {\adjustbox{margin=0.1em, bgcolor=c2}{gray}, respectively.

As seen from \pref{tab:friedman_ranking}, 
the algorithm selection systems using $A_{15}$ and $A_{10}$ perform the best  for $n=2$ and $n \geq 3$, respectively.
For any $n$, the algorithm selection system using $A_{50}$ shows the worst performance among the 10 algorithm selection systems.
This is because unlike the previous study~\cite{JankovicD20}, the actual maximum number of function evaluations for the whole algorithm selection system is exactly the same as $100n$.
In our study, an optimizer selected by the system can use only $100n - s$ function evaluations for the search.
Although a large $s$ is helpful to improve the quality of ELA features, it can degrade the performance of optimizers.
For any $s$, the algorithm selection systems using $A_{0}$ perform worse than those using $A_{10}$.
This result indicates the importance of constructing algorithm portfolios by considering the number of function evaluations actually available for optimizers.

\begin{table}[t]
\centering
\caption{Friedman test-based average rankings of the 10 algorithm selection systems for each dimension $n$.
}
\label{tab:friedman_ranking}
{\normalsize
\begin{tabular}{cccccccc}
\toprule
Portfolio, $s$ & $n=2$ & $n=3$ & $n=5$ & $n=10$\\
\midrule
$\set{A}_{0}, s=10n$ & 5.50 & 5.65 & 5.73 & 5.58\\
$\set{A}_{0}, s=15n$ & 5.54 & 5.48 & 6.02 & 6.06\\
$\set{A}_{0}, s=20n$ & 6.21 & 5.10 & 5.35 & 6.15\\
$\set{A}_{0}, s=25n$ & 5.50 & 4.85 & \cellcolor{c2}4.60 & 5.94\\
$\set{A}_{0}, s=50n$ & 5.00 & 5.85 & 6.00 & 5.98\\
$\set{A}_{10}, s=10n$ & \cellcolor{c2}4.44 & \cellcolor{c1}4.15 & \cellcolor{c1}4.12 & \cellcolor{c1}3.04\\
$\set{A}_{15}, s=15n$ & \cellcolor{c1}4.38 & \cellcolor{c2}4.52 & 5.37 & \cellcolor{c2}4.04\\
$\set{A}_{20}, s=20n$ & 5.29 & 5.65 & 5.79 & 4.67\\
$\set{A}_{25}, s=25n$ & 5.21 & 5.81 & 4.73 & 5.29\\
$\set{A}_{50}, s=50n$ & 7.94 & 7.94 & 7.27 & 8.25\\
\bottomrule
\end{tabular}}\\
\end{table}

\subsubsection{Comparison based on the prediction accuracy}

While the previous section focuses on the performance of algorithm selection systems, this section focuses on the prediction accuracy of algorithm selection systems.
Thus, this section discusses how accurately the best optimizer is selected from each algorithm portfolio.
Note that an algorithm selection system with high prediction accuracy does not necessarily perform well in the previous section.

\pref{fig:vbs_selected} shows the comparison of the 10 algorithm selection systems in terms of the prediction accuracy of the best optimizer on the 24 BBOB functions.
\pref{fig:vbs_selected} shows the percentage of the best optimizer selected by each algorithm selection system out of the 31 runs.
For $\set{A}_0$, we show the results using $s=25n$, which is the best setting in the previous section.

As seen from \pref{fig:vbs_selected}, the algorithm selection system using $\mathcal{A}_0$ shows the better performance than to that using $\mathcal{A}_{10}, \ldots, \mathcal{A}_{50}$ for $n = 2, 3, 5$.
We applied the Wilcoxon rank-sum test to the results and confirmed the significance in them.
Thus, we can say that the algorithm selection system using $\mathcal{A}_0$ can predict the best optimizer well.

Unlike $\set{A}_{10}, \ldots, \set{A}_{50}$, $\mathcal{A}_0$ is constructed based on the performance of each optimizer until $100n$ function evaluations.
In general, it is easy for optimizers to reach the optimal solution when the maximum number of function evaluations is set to large.
In addition, due to the property of the fixed-budget setting, there may be multiple best optimizers in algorithm portfolios.
Therefore, it is likely that $\mathcal{A}_0$ has more than one best optimizer on each BBOB function.
For this reason, it is easier for the algorithm selection system using $\set{A}_{0}$ to predict the best optimizer than that using $\mathcal{A}_{10}, \ldots, \mathcal{A}_{50}$.

However, as demonstrated in the previous section, the algorithm selection system using $\set{A}_{0}$ performs poorly in terms of the actual performance.
Although the algorithm selection system using $\set{A}_{0}$ has high prediction accuracy, there is a contradiction between the performance of optimizers.
The algorithm selection system using $\set{A}_{0}$ can select the best optimizer for $100n$ function evaluations, but the best optimizer for $100n$ function evaluations is not the best for $100n - s$ function evaluations.

\begin{figure}[t]
   \centering
\includegraphics[width=0.47\textwidth]{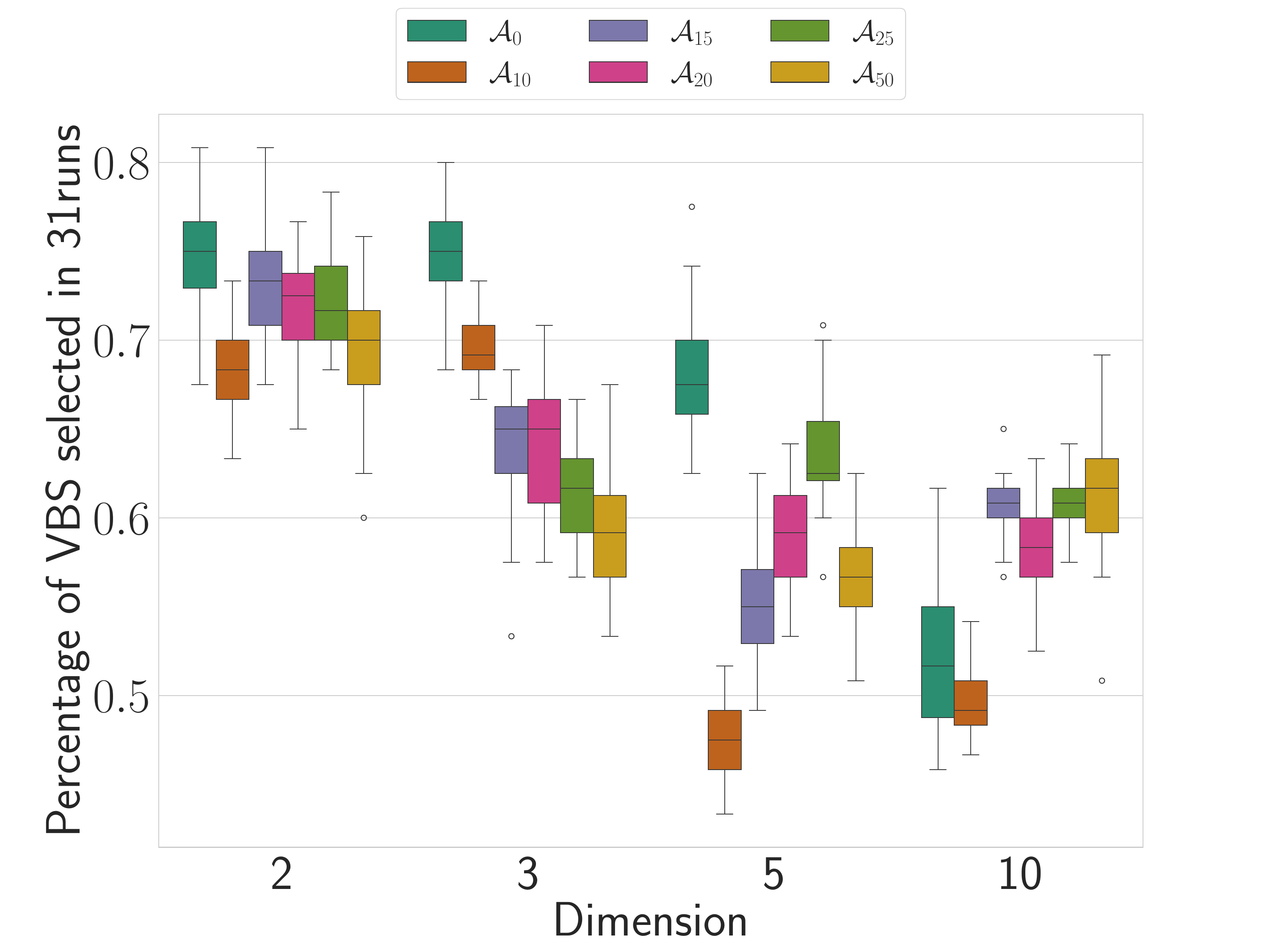}
\caption{Percentage of the best optimizer selected by each algorithm selection system out of the 31 runs, where higher is better.}
   \label{fig:vbs_selected}
\end{figure}

\begin{tcolorbox}[title=Answer to RQ1, sharpish corners, top=2pt, bottom=2pt, left=4pt, right=4pt, boxrule=0.5pt]
    Our results showed the importance of constructing algorithm portfolios by considering the number of function evaluations actually available for optimizers for computationally expensive optimization with the fixed-budget setting.
    Our results suggested using $s=10n$ when the maximum number of evaluations is $100n$.
    We also showed that the prediction accuracy of the best optimizer is not directly related to the performance of algorithm selection systems.
\end{tcolorbox}

\subsection{Comparison of algorithm selection systems and their SBSs}
\label{subsec:rq2}

\pref{tab:vs_sbs} shows pair-wise comparison of the algorithm selection system using each algorithm portfolio and its SBS on the BBOB function set for each dimension $n$.
\pref{tab:vs_sbs} shows the results of the algorithm selection systems using the six algorithm portfolios $\set{A}_0, \set{A}_{10}, \ldots, \set{A}_{50}$.
Here, the SBS for each algorithm portfolio is shown in \pref{tab:alg_portfolio}.
For $\set{A}_0$, we set $s=25n$ based on the results in \pref{subsec:rq1}.
We apply the Wilcoxon rank-sum test to the two results of the algorithm selection system using each algorithm portfolio and its SBS.
\pref{tab:vs_sbs} shows the sum of $+$, $-$, and $\approx$ on the 24 BBOB functions for each comparison.
Here, the symbols $+$ and $-$ indicate that an algorithm selection system using an algorithm portfolio $\set{A}$ performs significantly better ($+$) and significantly worse ($-$) than the SBS in $\set{A}$ according
to the Wilcoxon rank-sum test with $p < 0.05$.
The symbol $\approx$ means neither of them.
We highlight a result with {\adjustbox{margin=0.1em, bgcolor=c1}{dark gray} when an algorithm selection system using $\set{A}$ outperforms the SBS in $\set{A}$, i.e., the sum of $+$ is greater than the sum of $-$.

Note that the comparison with the SBS is relative.
For example, let us consider two algorithm selection systems $\texttt{AS}_1$ and $\texttt{AS}_2$ that use two algorithm portfolios $\set{A}_1$ and $\set{A}_2$, respectively.
Suppose that $\texttt{AS}_1$ performs better than the SBS in $\set{A}_1$, while $\texttt{AS}_2$ performs worse than the SBS in $\set{A}_2$.
In this case, one may conclude that $\texttt{AS}_1$ performs better than $\texttt{AS}_2$, but this is wrong.
Note also that it is difficult to outperform the SBS even in the common fixed-target setting~\cite{tanabe2022benchmarking} when using the LOFO-CV.
In addition, the difficulty of outperforming the SBS depends on algorithm portfolios~\cite{tanabe2022benchmarking}.

As shown in \pref{tab:vs_sbs}, the six algorithm selection systems are outperformed by their SBSs for $n=2, 3$.
In contrast, the algorithm selection systems using $\set{A}_{15}, \ldots, \set{A}_{50}$ outperform their SBSs for $n=5, 10$.
This result suggests that algorithm selection performs well as $n$ increases.

As shown in \pref{tab:vs_sbs}, only the algorithm selection system using $\set{A}_{10}$ fails to outperform its SBS for any $n$.
However, as $n$ increases, the difference between the sum of $+$ and $-$ becomes smaller.
In fact, the algorithm selection system using $\set{A}_{10}$ performs worse than its SBS only on one BBOB function for $n=10$.
As seen from \pref{tab:alg_portfolio}, the SBS for $\set{A}_{10}$ is DTS-CMA-ES, and the SBS for other algorithm portfolios is lq-CMA-ES.
%
%
This difference in the SBS is considered to be one of the reasons why only the algorithm selection system using $\set{A}_{10}$ is inferior to its SBS.
As mentioned in \pref{sec:setup}, the \texttt{nbc} feature class  is not available for $s=10n$ due to the too-small solution set size.
However, our preliminary experimental results showed that the existence of \texttt{nbc} does not significantly influence the performance of algorithm selection.


\begin{table}[t]
 \setlength{\tabcolsep}{2pt} 
\renewcommand{\arraystretch}{0.9}
  \centering
  \caption{Pairwise comparison of the algorithm selection system using each portfolio $\set{A}$ and the SBS in $\set{A}$ for each dimension $n$. The sum of the symbols $+/-/\approx$ on the 24 BBOB functions are shown.}
  \label{tab:vs_sbs}
{\normalsize
\begin{tabular}{cccccccc}
\toprule
 & $n=2$ & $n=3$ & $n=5$ & $n=10$\\
\midrule
$\set{A}_{0}, s=25n$ & $2/10/12$ & $6/11/7$ & \cellcolor{c1}$9/4/11$ & $6/7/11$\\
$\set{A}_{10}, s=10n$ & $0/8/16$ & $0/3/21$ & $0/4/20$ & $0/1/23$\\
$\set{A}_{15}, s=15n$ & $5/7/12$ & $8/12/4$ & \cellcolor{c1}$12/8/4$ & \cellcolor{c1}$13/3/8$\\
$\set{A}_{20}, s=20n$ & $7/9/8$ & $8/12/4$ & \cellcolor{c1}$11/8/5$ & \cellcolor{c1}$12/3/9$\\
$\set{A}_{25}, s=25n$ & $9/9/6$ & $8/12/4$ & \cellcolor{c1}$13/8/3$ & \cellcolor{c1}$12/3/9$\\
$\set{A}_{50}, s=50n$ & $3/7/14$ & $9/10/5$ & \cellcolor{c1}$8/4/12$ & \cellcolor{c1}$10/6/8$\\
\bottomrule
\end{tabular}
}
\end{table}

\begin{tcolorbox}[title=Answer to RQ2, sharpish corners, top=2pt, bottom=2pt, left=4pt, right=4pt, boxrule=0.5pt]

Our results showed that the algorithm selection systems using $\set{A}_{15}, \ldots, \set{A}_{50}$ outperform their SBSs for $n=5, 10$ for computationally expensive optimization with the fixed-budget setting.
Thus, it is expected that algorithm selection is useful as $n$ increases.

\end{tcolorbox}

\section{Conclusion}
\label{sec:conclusion}


This paper focused on the influence of algorithm portfolios on the performance of feature-based algorithm selection for computationally expensive optimization with the fixed-budget setting.
In our study, the maximum number of evaluations is limited to $100n$.
First, we pointed out some issues in the experimental setup of the previous study~\cite{JankovicD20} (\pref{sec:proposed_method}).
Then, based on the discussion, we presented a more practical experimental setup (\pref{subsection:our_study}).
Through analysis, we addressed two research questions described in \pref{sec:introduction}.
We demonstrated the importance of constructing algorithm portfolios by considering the number of function evaluations actually available for optimizers (\pref{subsec:rq1}).
We also demonstrated that algorithm selection systems using some algorithm portfolios can outperform their SBSs for $n=5, 10$ (\pref{subsec:rq2}).
 
Although we set the maximum number of function evaluations to $100n$, future work should investigate the influence of the maximum number of function evaluations on the effectiveness of algorithm selection.
We observed that the algorithm selection systems did not perform very well for the challenging LOFO-CV.
Thus, there is much room for improvement of algorithm selection systems.
As in \cite{BelkhirDSS16}, it may be promising to design new feature classes that are useful even when the size $s$ of the solution set $\set{X}$ is small.




\begin{acks}
  This work was supported by JSPS KAKENHI Grant Number \seqsplit{23H00491} and LEADER, MEXT, Japan.

\end{acks}

\bibliographystyle{ACM-Reference-Format}
\bibliography{reference} 










\end{document}